\documentclass[sigconf]{acmart}
\usepackage{booktabs} % For formal tables
\usepackage{algorithmic}
\usepackage[linesnumbered, ruled, vlined]{algorithm2e}
\usepackage{soul}
\usepackage{url}
\usepackage{graphicx}
\usepackage{amsmath}
\usepackage{bbm}
\usepackage{xcolor, soul}
\usepackage{natbib}
\usepackage{CJKutf8}
\usepackage{tabularx}
\usepackage{verbatim}
\usepackage{balance,flushend}

\usepackage[multiple]{footmisc}

\newcommand{\vpara}[1]{\vspace{0.05in}\noindent\textbf{#1 }}

\setcopyright{rightsretained}
\copyrightyear{2019} 
\acmYear{2019} 
\setcopyright{acmcopyright}
\acmConference[KDD '19]{The 25th ACM SIGKDD Conference on Knowledge Discovery and Data Mining}{August 4--8, 2019}{Anchorage, AK, USA}
\acmBooktitle{The 25th ACM SIGKDD Conference on Knowledge Discovery and Data Mining (KDD '19), August 4--8, 2019, Anchorage, AK, USA}
\acmPrice{15.00}
\acmDOI{10.1145/3292500.3330725}
\acmISBN{978-1-4503-6201-6/19/08}

\newcommand{\kobe}{\textit{KOBE}}

\begin{document}
% \title[Towards Attribute-Aware and Knowledge-Grounded Text Generation]{Towards Attribute-Aware and Knowledge-Grounded Text Generation in Online Recommendation}
% \title[]{Personalized and Informative Generation for Product Description in Online Recommendation}
\title[]{Towards Knowledge-Based Personalized Product Description Generation in E-commerce}
% \titlenote{Produces the permission block, and copyright information}
% \subtitlenote{The full version of the author's guide is available as
%   \texttt{acmart.pdf} document}

\author[Q. Chen*, J. Lin*, Y. Zhang, H. Yang, J. Zhou, J. Tang]{
    Qibin Chen$^{1*}$, Junyang Lin$^{2,3*}$, Yichang Zhang$^{3}$, Hongxia Yang$^{3\dagger}$, Jingren Zhou$^{3}$, Jie Tang$^{1\dagger}$
}
\affiliation{
    $^1$ Department of Computer Science and Technology, Tsinghua University
}
\affiliation{
    $^2$ School of Foreign Languages, Peking University
}
\affiliation{
    $^3$ Alibaba Group
}
\email{
  chen-qb15@mails.tsinghua.edu.cn, linjunyang@pku.edu.cn
}
\email{
  {yichang.zyc, yang.yhx, jingren.zhou}@alibaba-inc.com
}
\email{
   jietang@tsinghua.edu.cn
}

\renewcommand{\authors}{Qibin Chen, Junyang Lin, Yichang Zhang, Hongxia Yang, Jingren Zhou, Jie Tang}

% \authornote{Dr.~Trovato insisted his name be first.}
% \affiliation{%

% \authornote{The secretary disavows any knowledge of this author's actions.}
% \affiliation{%

% \authornote{This author is the
%   one who did all the really hard work.}
% \affiliation{%

% \renewcommand{\shortauthors}{}

\begin{abstract}
    Quality product descriptions are critical for providing competitive customer experience in 
    an e-commerce platform.
    % Taobao,\footnote{\url{https://www.taobao.com/}} the largest online E-commerce platform in China.
    An accurate and attractive
    %product 
    description not only helps customers make an informed decision but also improves the likelihood of purchase.
    %more attention to or buy certain products. 
    However, crafting a successful product description is tedious and highly time-consuming. 
    %work load can be extremely high for any E-commerce company and the 
    Due to its importance, automating the product description generation has attracted considerable interest from both research and industrial communities. 
    %attention. 
    %Traditional 
    Existing methods mainly use 
    %include 
    templates or statistical methods, and their performance could be
    %still far from satisfactory.
    rather limited.
    %which are inherently limited. 
    %Thus,
    %Taking a step towards this goal, this paper explores a new way to generate the description by combining the power of neural networks and knowledge base. 
    In this paper, we explore a new way to generate personalized product descriptions by combining the power of neural networks and knowledge base.
    Specifically, we propose a \textbf{K}n\textbf{O}wledge \textbf{B}ased p\textbf{E}rsonalized (or \kobe) product description generation model in the context of e-commerce. 
    %through neural networks. 
    %On the one hand, due to the success of self-attention in modeling text sequences,
    In \kobe,  we extend the 
    %effective 
    encoder-decoder framework, the Transformer, to a sequence modeling formulation using self-attention. 
    %On the other hand, 
    In order to make the description both informative and personalized, \kobe\ considers a variety of important factors during text generation, including product aspects, user categories, and knowledge base.
    %Furthermore, in this work, 
    %To provide a large dataset with ground-truth for fair comparison with different methods, 
    %we construct a dataset consisting of 2,129,187 descriptions annotated by manual. %humans as inputs. 
    % Very different from traditional text generation~\cite{xxx}
    % \jt{better have some citations here}
    % the generated knowledge in the E-commerce domain is required to be updated frequently, as it is highly connected to people's daily life. 
    % \zc{Need carefully reorg the following. As for me, I see a major contribution of multi-source information integration, including external knowledge, user implicit feedback (click), etc. However, the method is not new thus can be weakened. Not sure if the proposed conditional text generation (by word clustering) is well studied in the literature. }
    %Extensive e
    % deployment
    Experiments on real-world datasets demonstrate that the proposed method outperforms the baseline on various metrics.\footnote{Dataset and code available at \url{https://github.com/THUDM/KOBE}.}
    \kobe\ can achieve an improvement of 9.7\% over state-of-the-arts in terms of BLEU.
    We also present several case studies as the anecdotal evidence to further prove the effectiveness of the proposed approach.
    %Besides, our qualitative analyses reflect that our proposed method can change the focus of the generated description on different conditions and enables the description to reflect the relevant knowledge. \hongxia{What does the last sentence mean, quite confusing}
    % Furthermore, as we have clustered items into different topics with graph embeddings, we incorporate the topic information with the item description for better matching of users' interests
    The framework has been deployed in Taobao,\footnote{\url{https://www.taobao.com/}} the largest online e-commerce platform in China.
    % at \textit{Guess You Like}, the front page of Mobile Taobao.
    % Deployment is at \textit{Guess You Like}, the front page of the Mobile Taobao, a world leading E-Commerce company. 
\end{abstract}


% \ccsdesc[300]{Mathematics of computing~Graph algorithms}


\maketitle

\renewcommand{\thefootnote}{\fnsymbol{footnote}}
\footnotetext[1]{These authors contributed equally to this work.}
\footnotetext[2]{Corresponding Authors.}
\renewcommand{\thefootnote}{\arabic{footnote}}

% !TEX root = ./0.main.tex

\section{Introduction}

\begin{figure*}[t]
    \centering
    \includegraphics[width=\textwidth]{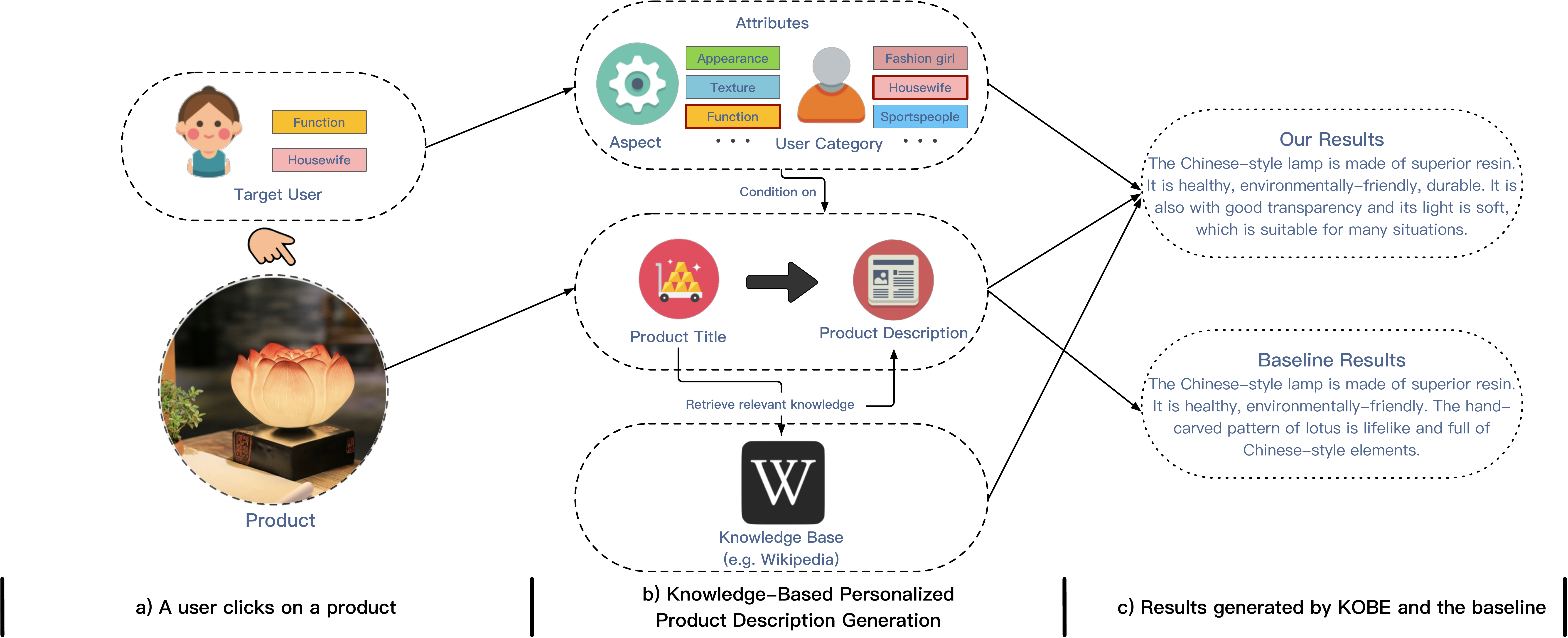}
    \caption{
        A motivating example of knowledge-based personalized product description generation.
        (a) A user clicks on a product, which is a Chinese-style resin lamp. The user focuses on the ``function'' product aspect and belongs to the ``housewife'' category.
        (b) The goal of \kobe\ is to generate a product description, given 1) the product title, 2) the desired product aspect and user category and 3) the relevant knowledge retrieved from an external knowledge base.
        (c) The first and second box displays the descriptions generated by our model and the baseline respectively.
        In this example, \kobe\ focus on the specified ``function'' product aspect and considers the user category ``housewife''.
        It also incorporates the knowledge ``resin is transparent''.
        In the meanwhile, the result produced by the baseline describes the product's appearance and style, which is less in this user's interest and might be more appropriate for younger user groups instead.
        The baseline result doesn't reflect external knowledge either.
        }
    % \caption{System demonstration in our real-world application.}
    % % \jt{the example is not that interesting. probably we can provide a procedure on how we generate the knowledge (with generation, knowledge reasoning, and dynamic updates}
    % \dm{Repaint the figure. No overlap. What does the question mean? And many other problems...}
    \label{fig:overview}
\end{figure*}

% \hide{
%%%%%%%%%%%%%%%%%%%%%%%%%%%%%%%%%%%%%%%%%%%%%%%%%%%%%%%%%%%%%%%%%%%%%%%%%%%%%%%%%%%%

% 简单陈述逻辑：
% 1）. 说明当前电商需要商品描述，因为：1. 人工批量生成有困难；2. 用户期待个性化和有内容的描述
% 2）. 说明当前的问题。第一，模板和统计规则的方法僵化死板，不能做到两点：1. 语言表述和话语结构不能像自然语言一样多样化；2. 无法个性化和体现知识
% 3）. 提出我们的方法：1. 我们第一个将seq2seq做到这个任务；2. 我们提出新模型，基于t2t拓展，包含两大方法：attribute fusion和knowledge incorporation，实现个性化和知识结合。
% 4.）. 提出一个新的数据集贡献给整个community

In e-commerce, product recommendation which aims at surfacing to users the right content at the right time~\cite{Christakopoulou:2018} plays an important role in providing a superior shopping experience to customers. 
% 这句话主要介绍推荐算法，感觉相关度不是那么大
% Traditional product recommendation relies on collaborative filtering~\cite{resnick1994grouplens,Sarwar:2001WWW} and recent advances in this field rely much on neural networks~\cite{Wang:2015KDD} to learn the representations of products and users.
One of the biggest challenges is to timely understand customers' intentions, help them find what they are looking for, and provide valuable assistance during their entire shopping process. %providing them product descriptions so that they can gain a comprehensive knowledge of the products.
% 这里想表达的逻辑：以上这些 product recommendation 算法都是做单品推荐，而我们认为单品推荐效果有限，因此线下门店会有 salespeople 主动过来根据顾客需要来 describe 商品的不同方面等，由此引出线上购物中商品描述的重要性。
Different from physical stores where salespeople could have a face-to-face conversation with customers, online stores in e-commerce heavily rely on textual product descriptions to provide crucial product information and, eventually, convince customers to buy the recommended products.
However, until now, most of the product descriptions in the online shopping platforms are still created manually, which is tedious, time-consuming, and less effective. On the one hand, textual descriptions should contain relevant knowledge and accurate product information to help customers make an informed decision. On the other hand, such descriptions should be personalized, based on customers' preferences, and promote their interests.
In this paper, we focus on automating the product description generation for online product recommendation. Specifically, the system can intelligently generate an accurate and attractive description for a specific product based on the product information and user preferences.

Recently, there has been increasing attention on automatic product description generation.
Most existing methods are based on templates and traditional statistical frameworks \citep{wang2017statistical, langkilde1998generation}.
% xxx proposed to use templates for the description generation, and \citep{wang2017statistical} applied statistical framework for the generation, which is able to project structured data of product to unstructured text.
Although applicable, these methods have some inherent drawbacks that they are limited in several respects.
They place a restriction on the phrasal expression and discourse structure, and they cannot generate personalized and informative descriptions. 
Moreover, they do not take any information about users or external knowledge into consideration.

Owing to the success of neural networks in natural language processing, we propose to apply neural methods and sequence-to-sequence (Seq2Seq) learning \citep{seq2seq} to product description generation. The Seq2Seq model has achieved tremendous success in natural language generation, including neural machine translation \citep{seq2seq, attention, GNMT} and abstractive summarization \citep{abs, IBM_sum}.  
% To the best of our knowledge, this is the first work that applies the neural methods and Seq2Seq model to this task.
% potential overclaim. There is existing work in the industry based on LSTMs.
Furthermore, as mentioned above, product descriptions should be personalized and informative. 
We therefore propose a \textbf{K}n\textbf{O}wledge \textbf{B}ased p\textbf{E}rsonalized (or \kobe) product description generation model for online recommendation. We extend the effective encoder-decoder framework, the self-attentive Transformer, to product description generation. Besides, two main components of \kobe, Attribute Fusion and Knowledge Incorporation, effectively introduce the product attributes and incorporate the relevant external knowledge. Figure~\ref{fig:overview} shows an example of knowledge-based personalized product description generation. 
% deployment
We also showcase how \kobe\ is deployed in the section \textit{Guess You Like}, the front page of Mobile Taobao.

To provide a large dataset for the automatic product description generation, we construct and make publicly available a new dataset of large volume, collected from Taobao.
The dataset contains 2,129,187 pairs of product basic information and descriptions created by shop owners from November 2013 to December 2018.
Each instance in the dataset is labeled with attributes.
For example, the user category attribute specifies the category of interest for potential readers.
We also retrieve the relevant knowledge about the products from a large-scale knowledge base in Chinese, CN-DBpedia \citep{xu2017cn}. 
Models trained on the dataset are able to learn how to generate descriptions based on titles, attributes and relevant knowledge. 

Our contributions are illustrated below:
\begin{itemize}
    \item We propose to extend neural methods and sequence-to-sequence learning to product description generation, and we implement our model based on the self-attentive Transformer framework. 
    
    \item To achieve effects in recommendation, we focus on the generation of personalized and informative product descriptions and propose a novel model \kobe\ that uses product attributes and external knowledge to improve the quality of generation. 
    
    \item We construct a new dataset of large volume, \textbf{TaoDescribe}, for product description generation. The dataset contains basic information of products and the corresponding descriptions. The data are collected from the real-world data in Taobao and we contribute this dataset to the community to nourish further development (refer to the Appendix).
    
    \item The evaluation and the analyses in the experiments demonstrate that our method outperforms the baseline model, and it is able to generate personalized and informative product descriptions.
\end{itemize}

\vpara{Organization}
The rest of the paper is organized as follows:
Section 2 formulates the knowledge-based personalized product description generation problem.
Section 3 introduces the proposed \kobe\ framework in detail.
In Section 4, we conduct extensive experiments and case studies.
Finally, Section 5 summarizes related work and Section 6 concludes this work.

% }

% !TEX root = ./0.main.tex

\section{Problem Formulation}
\label{sec:problem}

% \zc{The most attractive part to me is the definition of attribute and world facts, which makes me feel the contribution may highlight this.  }
In this section, we formulate the problem of automatic product description generation. 
% and further explain why it is possible to apply the sequence-to-sequence learning and our proposed methods to the problem.
%Basically, g
% Generating product description 
%at least 
% requires the basic information of the product, including its attributes.
% yet the internal conflict is that the source is 
The objective of the task is  
%data but the target is the unstructured text sequence. As our goal is 
to build a system that can generate 
product description automatically based on the input text.
% Thus to make the problem general enough, we formalize the problem as 
%we turn the original problem, which is ``structured data to text'', to a new problem,  which is 
% that of ``text to text'' generation (transformation). The details are illustrated below.
In the basic version of the problem, 
%in order to make the source data unstructured text, 
we take the product title as the input.
Given the product title represented as an input sequence of words $\boldsymbol{x} = (x_1, x_2, \ldots, x_n) \in \mathcal{X}$,
% with some specified attributes $\mathbf{a} \in \mathcal{A}$ (e.g. aspect, user category) and a large collection of world facts,
the objective of the system is to generate the description $\boldsymbol{y} = (y_1, y_2, \ldots, y_{m}) \in \mathcal{Y}$, a sequence of words describing the product. 
% in natural language. 
%The generated description $\boldsymbol{y}$ should approximate the target description $\boldsymbol{\hat{y}}$ in the training phase.
% , conditioned on these attributes and properly utilizing the world facts.
%\dm{what description? Longer? More natural? focus on aspect $a$?}
%
% Each product description is a sequence of words
% %\dm{better $\backslash$ textbf for vectors and sequence}
% $\mathbf{y} = (y_1, y_2, \ldots, y_{m}) \in \mathcal{Y}$.
% Note that in our training set, a product may have multiple reference descriptions, each having a different set of attributes.
%
Our ultimate goal is to obtain a personalized and informative description,
%As the goal of our study is the generation of 
%personalized and informative product description, 
so we introduce attributes and knowledge and further provide an improved definition.

\paragraph{Definition 2.1.}
\textbf{Attribute}
Each product title is annotated with multiple attribute %attributed
values.
% For $k$ attribute types,
There are $l$ attribute sets $\mathcal{A}_1, \ldots, \mathcal{A}_l$. For each attribute type, each product has its attribute value $a_i \in \mathcal{A}_i$. To generalize, it can be presented as $\boldsymbol{a} = (a_1, \ldots, a_{l}) \in \mathcal{A}_1 \times \cdots \times \mathcal{A}_l$.
% $\mathbf{a}$ represents attribute values $\mathbf{a} = $.
% Each attribute value $a_k$ is a discrete value in the set $\mathcal {A}_k$ of possible values for attribute $k$.
In our context, we have two attribute sets, where $\mathcal{A}_1$ represents the aspects of the products, such as quality and appearance, and $\mathcal{A}_2$ represents the user categories, which reflect the features of interested users.
$|\mathcal{A}_1|$ and $|\mathcal{A}_2|$ represents the number of aspects and the number of user categories.
% We have $|\mathcal{A}_1| = 3$ and $|\mathcal{A}_2| = 24$. %\jt{$\mathcal{A}_1$ is a set. right? why it is equal to 3?}
% For example, $\mathcal{A}_1 = \left\{fashion girl,housewife,sportspeople\right\}$ if $a_1$ represents the targeted user group of the description.
% \dm{Cannot understand. What is the difference between $\mathcal {A}_k$ and $\mathcal {A}$. Is there both aspect ``value'' and aspect? Tidy the concepts and notations up here. }

\paragraph{Definition 2.2.}
\textbf{Knowledge }
We consider knowledge as an extensive collection of raw text entries $\mathcal{W}$ (e.g., Wikipedia, CN-DBpedia) indexed by named entities in $\mathcal{V}$.
Each entry consists of a named entity $v \in \mathcal{V}$ as the key and knowledge $\boldsymbol{w} \in \mathcal{W}$ as the value, which is a sequence of words $\boldsymbol{w} = (w_1, w_2, \ldots, w_u)$. %\jt{symbols or words?}

% \vpara{Problem Attribute-Aware and Knowledge-Grounded Product Description Generation.}
% % \dm{What's that? product description generation. Do not generalize.}
% \dm{And rewrite the following.}
% Consider a training set $\mathcal { D } = \left\{\left( \mathbf{x}, \mathbf{a}, \mathbf{y} \right) \right\}$ and a collection of knowledge $\mathcal{W}$, our task is to learn a model $F : \mathcal{X} \times \mathcal{A} \times \mathcal{W} \rightarrow \mathcal{Y}$ that maps any pair $(\mathbf{x}, \mathbf{a}, \mathbf{w})$ of a product whose title is $\mathbf{x}$ and whose attributes are $\mathbf{a}$, plus a retrieved relevant world fact $\mathbf{w}$, to a description $\mathbf{y}$ that properly describe the product while conforming to the attribute $\mathbf{a}$. 
%\vspace{0.07in}
%Therefore, the problem can be defined as illustrated in the following: 
\vpara{Target problem:} Given the above definitions, our problem can be formally defined 
as generating a personalized and informative product description $\boldsymbol{y}$, 
based on the product title $\boldsymbol{x}$, the attributes $\boldsymbol{a}$ as well as the related knowledge $\boldsymbol{w}$.
%, the system is required to generate a corresponding product description $\boldsymbol{y}$.

% !TEX root = ./0.main.tex

\section{\kobe: Proposed Model Framework}

% The architecture we consider performs the mapping $ F : \mathcal { X } \times \mathcal { H } \times \mathcal { W } \rightarrow \mathcal { Y } $ through an encoder-decoder decoder framework.

\begin{figure}[t]
    \centering
    \includegraphics[width=0.42\textwidth]{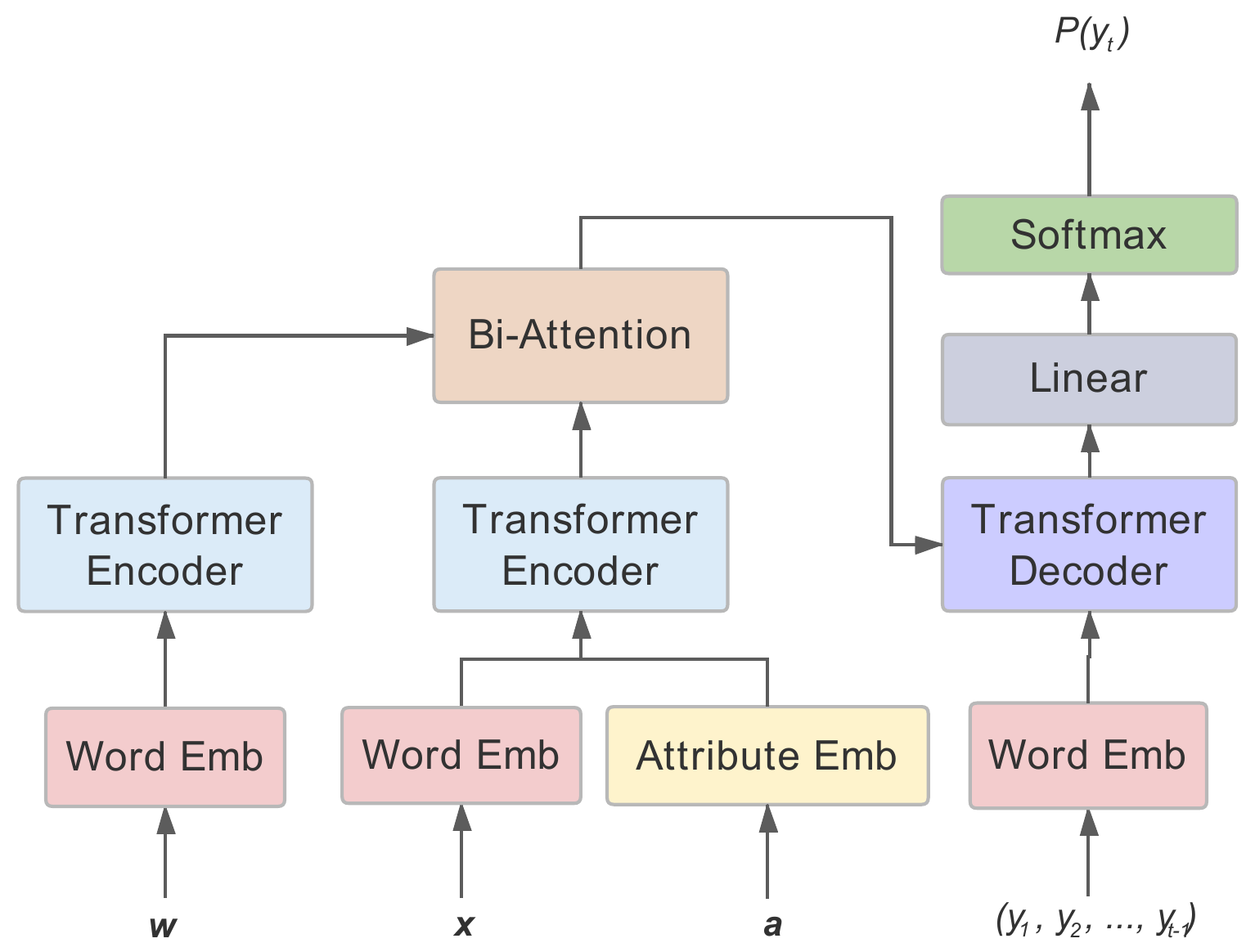}
    \caption{\kobe\ model architecture. We use six-layer transformer encoders for encoding the title $\boldsymbol{x}$ and the knowledge $\boldsymbol{w}$. The resulting representations $\boldsymbol{h}$ and $\boldsymbol{u}$ are then combined through bi-attention. We use a two-layer transformer decoder for decoding.}
    \label{fig:arch}
\end{figure}

In this section, we first review our basic framework Transformer~\citep{transformer}. 
% In our preliminary experiments, compared with the conventional RNN-based Seq2Seq \citep{seq2seq, attention}, Transformer demonstrates better performances in both effectiveness and efficiency. 
%Therefore, w
We then propose a novel model called \kobe. 
Besides the Transformer framework, \kobe\ consists of two other modules: Attribute Fusion and Knowledge Incorporation.
Attribute Fusion is responsible for integrating the attributes, including product aspects and corresponding user categories, with the title representation; and Knowledge Incorporation is responsible for incorporating the relevant knowledge retrieved from the knowledge base.
%Therefore, t
These two modules respectively contribute to the generation of personalized and informative product description.
As shown in Figure~\ref{fig:arch}, our proposed neural network model consists of a product title and attribute encoder, a knowledge encoder, a bi-attention layer, a description decoder, and an output layer.
In the following, we first introduce the basic framework of our model, namely Transformer, and then explain how we realize the personalization through the integration of attributes and how we improve the informativeness of the generation through the incorporation of knowledge in \kobe.

\subsection{An Encoder-decoder Framework}

% Recently, the Seq2Seq model based on RNN is the most common baseline framework. Yet, due to the problem of capturing long-term dependency, the RNN-based Seq2Seq has been improved with a number of sophisticated techniques but still suffers from low accuracy.
% More recently, researchers have been trying to replace RNN in the Seq2Seq framework and proposed two new frameworks, \textit{Convolutional Seq2Seq} which is based on convolutional neural networks \citep{fairseq} and 
\textit{Transformer} is based on vanilla feed-forward neural networks and attention mechanism \citep{transformer}.
In our implementations, the Transformer has demonstrated improved performances over the conventional frameworks, such as the RNN-based Seq2Seq. Thus
we focus on the details of Transformer in the following description.

\vpara{Encoder }
Generally, the encoder takes in the input text, and encodes it to a series of hidden representations. 
Specifically, the Transformer encoder takes in a sequence of words $\boldsymbol{x} = (x_1, x_2, \ldots, x_n)$, and sends it through an embedding layer to obtain the word embedding representations $\boldsymbol{e} = (e_1, e_2, \ldots, e_n)$. 
Besides the embedding layer, positional encoding is applied in order to represent positional information. We follow \citet{transformer} to use sinusoidal positional encoding. 
Then the word representations are transformed by the encoding layers to deep contextualized representations $\boldsymbol{h} = (h_1, h_2, \ldots, h_n)$.

On top of the embedding layer, the encoder is stacked with 6 identical layers, following \citet{transformer}. Take the first layer as an example. Inside the layer, the input representations are first transformed by multi-head self-attention. The attention score computation generates a distribution over the values $\mathbf{V}$ with the queries $\mathbf{Q}$ and keys $\mathbf{K}$, and the mechanism obtains an expectation of $\mathbf{V}$ following the distribution. 
% For example, in \citep{attention}, $\mathbf{Q}$ refers to the decoding state, and both $\mathbf{K}$ and $\mathbf{V}$ refer to the source context. 
The computation can be described below:
\begin{align}
    \mathbf{C} & = \alpha \mathbf{V} \label{context vector},       \\
    \alpha     & = \text{softmax}(f(\mathbf{Q, K})) \label{alpha},
\end{align}
\noindent where $\mathbf{C}$ refers to the output of attention, $\alpha$ refers to the attention distribution, and $f$ refers to the function for attention scores. 

As for self-attention, we first introduce the implementation of the uni-head attention and then the multi-head version. For the uni-head self-attention, the queries $\mathbf{Q}$, keys $\mathbf{K}$ and values $\mathbf{V}$ are the linear transformation of the input context $\boldsymbol{e}$. To be specific, $\mathbf{Q} = W_{Q}\boldsymbol{e}$, $\mathbf{K} = W_{K}\boldsymbol{e}$ and $\mathbf{V} = W_{V}\boldsymbol{e}$, where $W_{Q}$, $W_{K}$ and $W_{V}$ are weights. The representation of uni-head attention $\mathbf{C}_{self}$ can be presented as below:
\begin{align}
    \mathbf{C}_{self} = \text{softmax}\left(\frac{\mathbf{QK^{T}}}{\sqrt{d_k}}\right)\mathbf{V},
\end{align}
\noindent where $d_k$ refers to the size of each input representation $\boldsymbol{e}$.

Instead of using uni-head attention, our baseline model is equipped with multi-head attention, where we concatenate the attention representation of each head $\mathbf{C}_{self}^{(i)}$, $i \in \{1, \ldots, c\}$, where $c$ is the number of heads.
% \footnote{$c$ is the number of heads. We empirically set it as 8 in our setting.}
Specifically, for each head, we use the same computation for uni-head attention to generate the attention representation of the $i$-th head $\mathbf{C}_{self}^{(i)}$.
% $\mathbf{Q^{(i)}} = W_{Q}^{(i)}\boldsymbol{e}$, $\mathbf{K^{(i)}} = W_{K}^{(i)}\boldsymbol{e}$ and $\mathbf{V^{(i)}} = W_{V}^{(i)}\boldsymbol{e}$, and the representation of each head $\mathbf{C}_{self}^{(i)}$ and the final representation $\mathbf{C}_{self}$ are presented as below:
% %\jt{why some superscript is $^{(i)}$ while the other is $^i$. make it consistent}
% \begin{align}
%     \mathbf{C}_{self}^{(i)} & = \text{softmax}\left(\frac{\mathbf{Q^{(i)}K^{(i)T}}}{\sqrt{d_k}}\right)\mathbf{V^{(i)}} ,     \\
%     \mathbf{C}_{self}     & = [\mathbf{C}_{self}^{(1)}, \ldots, \mathbf{C}_{self}^{(i)}, \ldots, \mathbf{C}_{self}^{(m)}].
% \end{align}
In the next step, the output representations are sent into the Point-Wise Feed-Forward Neural Network (FFN), whose computation can be described in the following:
\begin{align}
    FFN(z) = W_2(ReLU(W_1z + b_1)) + b_2
\end{align}
where $z$ refers to the input of FFN.
% Moved to appendix
% Following \cite{transformer}, we empirically set the size of the inner representation ($d_{ff}=2048$) to four times of $\boldsymbol{e}$'s size ($d_{model}=512$). Except for the two main components, the other units in an encoder layer include residual connection \citep{resnet} and layer normalization \citep{LayerNorm}.

\vpara{Decoder }
Similar to the encoder, the decoder is stacked with 2 decoding layers containing multi-head self-attention and FFN. 
% In the detailed implementation for multi-head self-attention for the decoder, the attention does not attend to the subsequent positions, making it possible for the decoder to receive the whole ground-truth text in a non-autoregressive fashion. 
% Such masking for attention is identical to the attention at the inference stage, which generates outputs autoregressively and only attends to the previous positions. 
Furthermore, the decoder of Transformer performs attention on the source-side context, which is the final representation of the encoder $\mathbf{h}$. Here, the attention is named ``context attention'' for the distinction from ``self-attention''.  Context attention is still the aforementioned multi-head attention. The query $\mathbf{Q}$ is the linear transformation of the decoder state, and the keys $\mathbf{K}$ and values $\mathbf{V}$ and the linear transformation of the source-side context.

\vpara{Training }
The training of Transformer is based on the idea of maximum likelihood estimation. The objective is to generate a sequence to approximate the \textit{target} based on the input sequence. Formally, we have:% which can be presented as in the following:
\begin{align}
    \label{eq:seq2seq}
    P ( \boldsymbol { y } | \boldsymbol { x } ) = \prod _ { t = 1 } ^ { m } P \left( y _ { t } | y _ { 1 } , y _ { 2 } , \ldots y _ { t - 1 } , \boldsymbol { x } \right),
\end{align}
\noindent where $\boldsymbol{x}$ refers to the source sequence, $\boldsymbol{y}$ refers to the target sequence, and $t$ refers to the decoding time step.

To put into a nutshell, our proposed model is based on the self-attentive Transformer, which essentially is an encoder-decoder framework. We utilize the advantage of self-attention
% to adapt to 
in our task to generate product descriptions.
In Section~\ref{sec:attribute}~and~\ref{sec:knowledge}, we will describe how our proposed method enables personalization and informativeness for product description generation.
% \hongxia{which consists of two parts: attribute aware and knowledge-based. }
%%%%%%%%%%%%%%%%%%%%%%%%%%%%%%%%%%%%%%%%%%%%%%%%%%%%%%

% \subsubsection{Learning Objective.}
% The encoder-decoder framework can be trained by maximizing the log probability of a description $\boldsymbol{y}$ given the product title $x$.
% Then the training objective is as follows:

% \begin{align}
%     J\left(\theta\right) = - \sum _ { t = 1 } ^ { m } \log p \left( y _ { t } | y _ { 1 } , \ldots , y _ { t - 1 } , x ; \theta \right)
% \end{align}

% \subsection{\kobe} \label{sec:model}

% In this section, we introduce the details of our proposed method \kobe\ and explain how it can contribute to the improvement of personalization and informativeness in product description generation.

% The details are described in the following.

\subsection{Attribute Fusion} \label{sec:attribute}

One limitation of the baseline encoder-decoder framework is that it often gives general and vague descriptions which are often boring and useless.
In addition, it does not consider the 
%characteristics
preferences of the users
 %of potential reader groups 
 and usually ``speak'' in a  monotonous flavor.
To alleviate these 
%two 
problems, we not only consider product titles and descriptions but also intake ``flavored'' specific attributes, such as aspects and user categories.
We hope to generate different ``flavored" descriptions targeting at different groups of people, for example, formal 
%and official 
style for office people, fashion styles for adolescents, and so forth. Following \citet{ficler2017controlling, sennrich2016controlling}, we obtain these attributes from two sources: (1) the description using a heuristic and (2) user feedback associated with the description.
% This simulates the setting in which we specify the attributes (e.g. targeted user group) to the model and generate the ``flavored" descriptions accordingly.

\vpara{Aspects }
% The aspects are extracted from the descriptions using a heuristic.
A product description may contain information about different aspects of the product. Through a preliminary study on the large dataset (Cf.~Section~\ref{sec:experiments} for the detail of the dataset), we found that a successful description usually 
%In \citep{wu2017inferring}, the authors match texts to nine existing emotion categories. Inspired by their work, we observe that instead of expressing a kind of subjective emotions, the product descriptions are often written in a more objective way, and they often have
has a ``focus'' on one aspect of the product. 
%An aspect usually represents the perspective to view a product. 
For example, for a lamp, the focused aspect might be ``appearance'' with words like ``beautiful'' and ``delicate''  describing its external features.
%, a description for a vase, which pays attention to this aspect, usually describes its external features such as pattern with words like ``beautiful'' and ``delicate''.
Based on our statistics of the dataset, we empirically select $|\mathcal{A}_1|$ aspects, e.g., ``appearance'', ``function''.

Due to the massive scale of our data set, precisely labeling the aspect of each description in the dataset is not possible.
We thus extract the aspect from the description using a heuristic method based on semantic similarity.
Each description $\boldsymbol{y}\in\mathcal{Y}$ is labeled with an aspect $a_1$, based on the semantic similarity between the description and the aspects.
Details about the choice of the aspects and labeling methods are introduced in the additional pages (Cf.~Appendix~\ref{sec:appendix}).

% Table \ref{tab:aspects} shows examples and the number of adjectives for every aspect.
%
% We then automatically label each description $\boldsymbol{y}\in\mathcal{Y}$ in the dataset with an aspect $a_1$, based on the semantic similarity between the description and the set of adjectives belonging to the aspect.

\begin{figure}[t]
    \centering
    \includegraphics[width=0.42\textwidth]{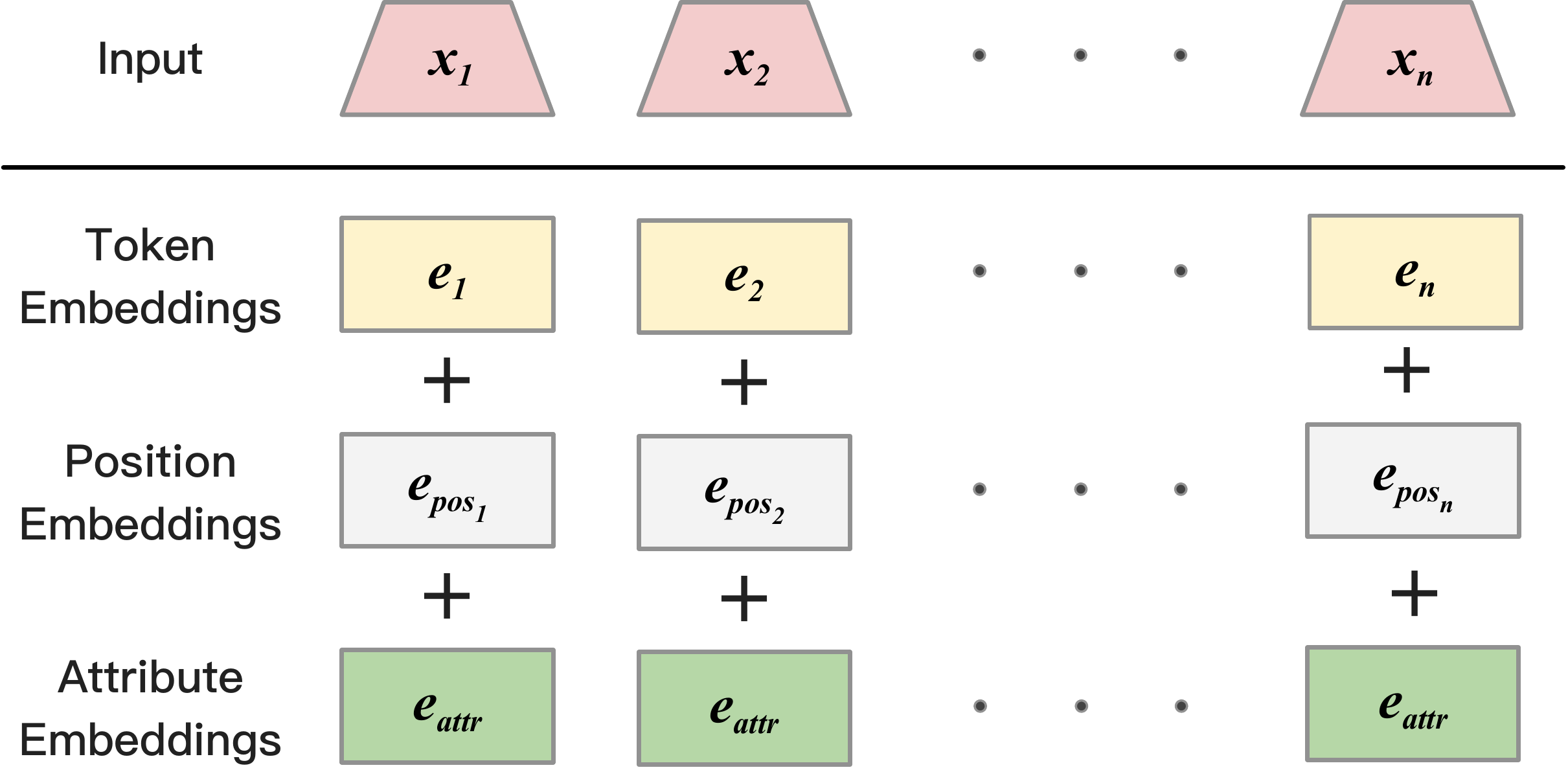}
    \caption{The input embeddings of the conditioned model are the sum of the title token embeddings, the position embeddings and the attribute embeddings.}
    \label{fig:input}
\end{figure}

\vpara{User Categories }
%Apart from discovering different 
Besides the different aspects of each description, we also try to discover the group of users who would be interested in a description.
% will interest, so that we can 
This is important to further design personalized description
%manipulate the description 
for target users. In our e-commerce setting, each user is labeled with ``interest tags'', which derive from his/her basic information, e.g., browsing history and shopping records.
In our context, we consider the user's most important tag as his/her category and each user is assigned to one of the $|\mathcal{A}_2|$ categories.

In the ideal conditions, we have users' implicit feedback (click, dwell time) data for each description $\boldsymbol{y}$ in the dataset.
%Naturally, e
Each description can then be assigned to a user category, according to the major group of users that have clicked or stayed long enough on this description.
%\footnote{
Note that soft assignment to user categories is also feasible in our framework. We find that hard assignment is simple and effective enough and thus we use hard assignment in our experiments and online systems.

%However, we observe that only a small portion of 
We find that user feedbacks on the product descriptions in our dataset are very sparse.
%have enough user feedback.
This results in a lot of noise in the assigned user categories. 
To overcome this problem, we collect another set of texts $\mathcal{Z}$ in our e-commerce context.
These texts are similar to product descriptions but have much more user feedback.
% \jt{no need to say ``due to'', simply clarify how the text has been collected}
Then we train a CNN-based \citep{kim2014convolutional} multi-class classifier on $\mathcal{Z}$ which takes the text as input and predicts the user category it belongs to.
% The classifier achieves an accuracy of 81.0\% on the validation set.
After that, 
%Then
we use the trained classifier $M$ to label $\mathcal{Y}$ and obtain a user category attribute $a_2$ for each $\boldsymbol{y} \in \mathcal{Y}$.
The collection of ``interest tags'', the dataset $\mathcal{Z}$, the architecture of the classifier and training procedure will be described with details in the additional pages (Cf.~Appendix~\ref{sec:appendix}).

\vpara{Attribute Fusion Model }
In a conditioned encoder-decoder framework similar to \citep{ficler2017controlling, tang2016context}, we add an additional conditioning context, which is attribute $\boldsymbol{a}$ in our case.
Then Equation~\ref{eq:seq2seq} is changed to:%into the following:
\begin{align}
    P \left( \boldsymbol{y} | \boldsymbol{x}, \boldsymbol{a} \right) = \prod _ { t = 1 } ^ { m } P \left( y _ { t } | y _ { 1 } , y_{2}, \ldots y _ { t - 1 } , \boldsymbol{x}, \boldsymbol{a} \right).
\end{align}
There are several ideas to condition the model on a context \citep{sennrich2016controlling,michel2018extreme,li2016persona}.
However, most of them are initially designed for LSTMs.
For Transformer, we experimented with all these methods (Cf.~Section~\ref{sec:experiments}).
Our results show that it achieves the best performance by simply adding the attribute embedding to the title embeddings.
% outperforms all these.

Specifically, we first embed the aspect attribute $a_1$ and user category attribute $a_2$ to obtain the attribute representation $\boldsymbol{e}_{a_1}$ and $\boldsymbol{e}_{a_2}$.
We average the embedding of the two attributes to obtain a single representation $\boldsymbol{e}_{attr}$.
Then, given the product title embeddings $\boldsymbol{e} = (e_1, e_2, \ldots, e_n)$, we add the attribute embedding to the title word embedding $e_i$ at each timestamp $i$.
The fused representation is illustrated in Figure~\ref{fig:input}. A similar idea has also been used in BERT~\cite{BERT}.

% We demonstrate that the aspects and user tags, though are collected using different methods, both work effectively. 
% In our experiments
In Section~\ref{sec:experiments}, we will demonstrate that both the models with the aspects and user categories work effectively.
In addition, the attribute fusion model is straightforward to generalize to other kinds of attributes.

% \begin{align}
%     \tilde{e}_i = e_i + \boldsymbol{e}_{attr}
% \end{align}

\subsection{Knowledge Incorporation}\label{sec:knowledge}
% The problem with only using the product title as input is that the title itself can be heavily optimized by SEO techniques or manually devised by the shop owner and thus consists of fancy and redundant words.
% We observe that product titles themselves are often noisy and lack of necessary information for generating the right description.
% However, humans are able to write a description for the product merely by just looking at the title.
%In order to achieve the effect of informativeness, 
The basic product information
%, including product title and attributes, 
is still far from enough for generating an interesting and informative product description, for the reason that in the real-world context, humans generate descriptions not only based on the product information, but also their commonsense knowledge in mind about the product. 
For example, when generating the description for a resin lamp, one 
%cannot write a description only with the title, but instead one shall 
may use his/her knowledge, such as the features of resin, to compose an informative description. 
%Similarly, 
Inspired by this, we %hypothesize 
%that our model
propose that the generation
should make a good combination of the basic information and the relevant external knowledge.
% so as to reach the aforementioned goal. 
We thus consider grounding the basic information of the specified products with the associated knowledge \citep{ghazvininejad_knowledge-grounded_2017},  
%with the purpose of simulating 
following the procedure where people utilize their knowledge to describe the product.

To be more specific, we first retrieve the relevant knowledge with the product information from the knowledge base and then encode the knowledge, which is separated from the encoding of the product title.
Before entering the phase of decoding, both encoded representations are integrated with our proposed techniques. Thus the decoding can be based on both the basic information and the relevant knowledge.
We discuss our proposed method for knowledge incorporation, including knowledge retrieval, knowledge encoding as well as the combination, in the following.

\vpara{Knowledge Retrieval }
Knowledge Retrieval is responsible for retrieving relevant knowledge for a product from the knowledge base $\mathcal{W}$.
Specifically, we obtain the relevant knowledge for our input products from CN-DBpedia \citep{xu2017cn}, which is a large-scale structural knowledge graph in Chinese. The dataset that we construct is in Chinese to be introduced in detail in Section~\ref{sec:dataset}, and CN-DBpedia perfectly matches the requirements of our purpose.
As we assume that the relevant knowledge of each word in the product title should be relevant to the product itself, the words in the product title $\boldsymbol{x}$ are the search items for the relevant knowledge.
Formally, given a product title $\boldsymbol{x} = (x_1, x_2, \ldots, x_n)$, we match each word $x_i$ to a named entity $v_i \in \mathcal{V}$, which should be a vertex in the knowledge graph. Then we retrieve the corresponding knowledge $\boldsymbol{w}_i$ from $\mathcal{W}$ based on the named entity $v_i$.
Furthermore, as there are multiple knowledge concepts describing $v_i$ in the knowledge base, we randomly sample 5 candidates as its matched knowledge for each product.
We then concatenate the knowledge according to their original order (i.e., the order in which their corresponding named entity appears in the product title), separated by a special token <SEP>.

\vpara{Knowledge Encoding and Combination }
%Since we believe that t
The retrieved knowledge and the basic information are somehow different, describing different aspects of the product. A proper mechanism should be designed to combine the two pieces of information.
%of diverse aspects, they should not be encoded together but instead, the knowledge should be encoded separately and a proper mechanism should be implemented for an effective combination.
We set a \textbf{Knowledge Encoder}, which is a self-attentive Transformer encoder.
It is responsible for encoding the retrieved knowledge to a high-level representation $\boldsymbol{u}$. 
The obtained representation $\boldsymbol{u}$ is similar to the encoded representation $\boldsymbol{h}$ of the product title $\boldsymbol{x}$.
%Though similar, the two representations are encoded by different encoders.
%Simple addition may cause a negative impact by the noise in the texts for the product title.
%Then 
We then apply 
the BiDAF (bidirectional attention flow \citep{seo2016bidirectional}) to combine the two kinds of representations.
%is applied to combine the title and fact representations.
% Formally, given the encoded title representation $\boldsymbol{h}$ and its world facts representation $\boldsymbol{h}_w$. The combined representation is computed as follows. 

More specifically, we compute attentions in two directions: title-to-knowledge attention and knowledge-to-title attention.
Both of these attentions are derived from a shared similarity matrix, $\mathbf { S } \in \mathbb { R } ^ { n \times u }$, between the contextualized representations 
% $\boldsymbol { h } = \left( h _ { 1 } , h _ { 2 } , \ldots , h _ { n } \right), 
$\boldsymbol{h} \in \mathbb{R}^{n \times d}$ of the title $\boldsymbol{x}$ and 
% $\boldsymbol { u } = \left( u _ { 1 } , u _ { 2 } , \ldots , u _ { u } \right)$, 
$\boldsymbol{u} \in \mathbb{R}^{u \times d}$ of the knowledge $\boldsymbol{w}$. $\mathbf { S } _ { i j }$ indicates the similarity between the $i$-th title word and the $j$-th knowledge word.
The similarity matrix is computed as:
\begin{align}
    \mathbf { S } _ { i j } = \alpha \left( h _ { i } , u _ { j } \right) \in \mathbb { R },
\end{align}
\noindent where $\alpha$ represents the function that encodes the similarity between the two input vectors.
We choose $\alpha ( h , u ) = \mathbf { w } _ {  \mathbf { S } } ^ { \top } [ h ; u ; h \circ u ]$, where $\mathbf { w } _ { \mathbf { S } } \in \mathbb { R } ^ { 3 d }$ is a trainable weight vector ($d$ is the dimension of contextualized representation) , $\circ$ is element-wise multiplication, $\left[; \right]$ is vector concatenation across row, and implicit multiplication is matrix multiplication.
Now we use $\mathbf{S}$ to obtain the attentions and the attended vectors in both directions.

Title-to-knowledge attention signifies which knowledge words are most relevant to each title word.
Let $\mathbf { a } _ { i } \in \mathbb { R } ^ { u }$ represent the attention weights on the knowledge words by the $i$-th title word, $\sum _ {j} \mathbf { a } _ { i j } = 1$ for all $i$.
The attention weight is computed by $\mathbf{ a }_{ i } = \operatorname { softmax } \left( \mathbf { S } _ { i : } \right) $, and subsequently each attended knowledge vector is $\tilde { u } _ { i } = \sum _ { k } \mathbf { a } _ { i k } u _ { k }$.
Hence $\tilde { \boldsymbol{u} } \in \mathbb{R} ^ {n \times d}$ contains the attended knowledge vectors for the entire title.

Knowledge-to-title attention signifies which title words have the closest similarity to one of the knowledge words.
We obtain the attention weights on the title words by $\mathbf { b } = \operatorname { softmax } \left( \max ( \mathbf { S } _ {i :} ) \right)$.
% \in \mathbb { R } ^ { n }$.
Then the attended title vector is $\tilde { h } = \sum _ { k } \mathbf{b}_{k} h_{k}$.
% \in \mathbb { R } ^ { d }$
This vector indicates the weighted sum of the most important words in the title with respect to the knowledge.
$\tilde { h }$ is tiled $n$ times across the column, thus giving $\tilde { \boldsymbol { h } } \in \mathbb { R } ^ { n \times d }$.
Similar to  \citep{seo2016bidirectional}, we use a simple concatenation $[ \boldsymbol { h } ; \tilde { \boldsymbol { u } } ; \boldsymbol { h } \circ \tilde { \boldsymbol { u } } ; \boldsymbol { h } \circ \tilde { \boldsymbol { h } } ] \in \mathbb { R } ^ { 4 d \times T }$ to get the final combined representation.
% \begin{align}
%     \label{eq:biatten}
%     \hat { \boldsymbol{o} } = BiAttention(\boldsymbol{o}, \boldsymbol{o}_w)
% \end{align}

% We also considered alternatives to BiDAF
%this bi-attentional way of 
% to combine the title and the retrieved knowledge, such as concatenation and summation. However, BiDAF yields the best results in all evaluation metrics.

% !TEX root = ./0.main.tex

\section{Experiments} \label{sec:experiments}
In this section, we conduct extensive experiments to evaluate the performance of the proposed model.
We first introduce the dataset, the compared baseline models and the evaluation metrics.
We also demonstrate the experimental results in a series of evaluations and perform further analyses on the effectiveness of our approach in generating both personalized and informative descriptions.
Details about the experiment setup and implementation will be described in the additional pages (Cf.~Appendix~\ref{sec:appendix}).

\subsection{Dataset} \label{sec:dataset}
In order to further the study of product description generation, considering that there is a lack of large-scale dataset for this task, we constructed a new dataset \textbf{TaoDescribe}, containing the basic information of the products, including the title, the aspect and the user category, as well as the product description.
The data are collected from Taobao, a large-scale website for e-commerce in China.
% In brief, the dataset consists of a large number of instances, each a pair of product information and product description.

% We collected the aforementioned pairs from the data in Taobao, which dates from November 2013 to December 2018.
The product information and the description are composed by the sellers and content producers on the website from November 2013 to December 2018.
Each data instance is automatically annotated with its aspect and user category using the methods introduced in Section~\ref{sec:attribute}.
Each data instance is also matched with knowledge retrieved from the knowledge base CN-DBpedia.
This enables product description generation to make use of such information for personalization and knowledge incorporation.
Overall, the dataset contains 2,129,187 $(\mathbf{x}, \mathbf{y}, \mathbf{a}, \mathbf{w})$ instances after preprocessing.
% Furthermore, we observe that the samples are noisier if the product title or the product description is too long. Therefore, we discarded pairs with a title longer than 100 tokens or with a description longer than 150 tokens.
More details about the dataset will be described in the additional pages (Cf.~Appendix~\ref{sec:appendix}).

\subsection{Systems for Comparison}
In this section, we introduce the baseline and choices for our model components.
 
\vpara{Attribute Fusion }
We first investigate common choices of conditioning a sequence-to-sequence model on input attributes.

\begin{itemize}
    \item \textbf{Baseline }
          Our baseline is a Transformer, with the product title as its source sequence and the description as its target sequence, without considering attributes or incorporating knowledge.
    \item \textbf{Attr-D} (\textbf{D}edicated)
          Before moving on to conditioned models, we train a dedicated model on each subset of the data following \citet{ficler2017controlling}.
          For example, for three aspects ``appearance'', ``texture'' and ``function'', we split the dataset into three subsets and train an unconditional model on each subset.
          We test them on the corresponding portions of the test set and average the scores of all testing instances.
          However, this method does not scale to multiple attributes or attributes with more possible values (e.g., user category) because there will be too many models to be trained and the dataset for each model will be tiny. Therefore, We train dedicated models for the aspects only.
    \item \textbf{Attr-S} (\textbf{S}ource token)
          Following \citet{sennrich2016controlling}, we can add attributes as special tokens to the source text (i.e., product title). In our experiments, we add one of <A-1>, <A-2>, ..., <A-$|\mathcal{A}_1|$> to represent the aspect and one of <U-1>, <U-2>, ... , <U-$|\mathcal{A}_2|$> to represent the user category.
    \item \textbf{Attr-T} (start of \textbf{T}arget sequence)
          Another technique to attribute fusion is to use an attribute-specific start token for the target sequence \citep{michel2018extreme}, instead of a shared token.
          For multiple attributes, we first embed each attribute value and then feed the averaged embedding as the start of sequence token embedding following \citet{lample2018multipleattribute}.
    \item \textbf{Attr-A} (\textbf{A}dd)
          We simply add the attribute embeddings to the title embeddings at each timestep as introduced in Section~\ref{sec:attribute}.
        %   At each timestep during encoding, we embed the attribute and concatenate the attribute embedding to the input embedding and we project the combined embedding to the original embedding size.
          For multiple attributes, we embed them separately and average the attribute embeddings before adding to the title embeddings.
\end{itemize}

We conducted three experiments with each Attribute Fusion model: 1) only adding aspects, 2) only adding user categories, 3) adding both attributes. They are denoted as Attr-S (Aspect), Attr-S (User), Attr-S (Both).

\vpara{Knowledge Incorporation }
We then compare the effect of incorporating knowledge using bi-attention.% in Equation \ref{eq:biatten}.

\begin{itemize}
    % \item \textbf{Know-A} (\textbf{A}dd)
    %       We use a method similar to \citep{ghazvininejad_knowledge-grounded_2017} that adds the knowledge encoding to the source encoding.
    %       \begin{align}
    %           \hat { \boldsymbol { h } } =  \boldsymbol { h } + \boldsymbol { u }
    %       \end{align}
    % \item \textbf{Know-C} (\textbf{C}oncat)
    %       We also try concatenating the knowledge encoding with the source embedding and then project them to the embedding size.
    %       \begin{align}
    %           \hat { \boldsymbol { h } } = \mathbf{W} \left[ \boldsymbol { h }; \boldsymbol { u } \right]
    %       \end{align}
    %       \noindent where $\mathbf{W} \in \Real^{2d\times d}$, $d$ is the embedding dimension size; $\left[;\right]$ is vector concatenation.
    % \item \textbf{Know-UniAttn}
    \item \textbf{Know-BiAttn}
          % (\textbf{Bi}-\textbf{D}irectional \textbf{A}ttention \textbf{F}low) \citep{seo2016bidirectional}
          %   Compared to the above methods, bi-attention
          % %   introduced in \citep{seo2016bidirectional} 
          %   has been very successfully applied to question answering (QA) tasks. We thus adopt this model to help to model more complex interactions between the title and knowledge.
          Know-BiAttn refers to the model equipped with bidirectional attention,  which aims at integrating the title representation and knowledge representation as introduced in Section~\ref{sec:knowledge}.
\end{itemize}

\vpara{\kobe\ }
Our final model \kobe\ combines \textbf{Attr-A (Both)} and \textbf{Know-BiAttn}.
\subsection{Evaluation Metrics}
We evaluate our model on generation quality, diversity and attribute capturing ability.

\vpara{BLEU }
We first verify that the introduction of the attributes indeed helps in achieving better results in terms of the BLEU score as a sanity check \citep{bleu}.
We compare the test BLEU score of our conditioned model to the unconditioned baseline trained on the same data.

\vpara{Lexical Diversity }
A common problem in automatic text generation is that the system tends to generate safe answers without enough diversity \citep{jiwei_adversarial, jiwei_rl}. Thus we also evaluate the lexical diversity to observe whether our model is able to generate diversified texts. This is also a reflection of informativeness as human-generated texts are usually full of diversified contents.
A low diversity score often means generated contents are general and vague, while higher diversity means the generated contents are more informative and interesting. We calculate the number of distinct n-grams produced on the test set as the measurement of the diversity of generated descriptions.
% We denote the set of descriptions generated by a model on the test set as $\hat{Y}_T$. $G_n$ is a function that maps a sequence of tokens to the set of its n-grams.
% Then the n-gram diversity score is calculated as follows:
% \begin{align}
% |\cup_{\boldsymbol{\hat{y}} \in \hat{Y}_T} G_n(\boldsymbol{\hat{y}})|
% \end{align}
% \begin{figure}[t]
%     \centering
%     \includegraphics[width=0.48\textwidth]{figure/diversity.pdf}
%     \caption{Lexical Diversity}
%     \label{fig:diversity}
% \end{figure}

\vpara{Attribute Capturing }
However, BLEU and lexical diversity are not sufficient for measuring how well the generated results correlate to the specified attributes.
In particular, we are interested in evaluating how difficult it is to recover the input attributes from the descriptions generated by a model.

For aspects, it is straightforward to run the automatic tagging procedure on our generated descriptions; for user categories, we use the pretrained classifier to predict the user category of each generated description.
For attribute $k$, the attribute capturing score is computed as the prediction accuracy:
\begin{align}
    \label{eq:accuracy}
    \frac{1}{|\hat{Y}_T|} \sum _ {(\boldsymbol{\hat{y}}, \boldsymbol{y}) \in (\hat{Y}_T, Y_T)} \mathbbm{1} _ {  M(\boldsymbol{\hat{y}}) = M(\boldsymbol{y}) },
\end{align}
\noindent
where $M: \mathcal{Y} \rightarrow \mathcal{A}_2$ denotes the user category classifier and $\mathbbm{1} _ {  M(\boldsymbol{\hat{y}}) = M(\boldsymbol{y}) }$ is the indicator function with value as 1 when $M(\boldsymbol{\hat{y}}) = M(\boldsymbol{y})$ and 0 otherwise.

\subsection{Result Analysis}
In this section, we conduct an analysis of our proposed model to evaluate the contribution of attributes and knowledge. We focus on a few issues illustrated below.

% \vpara{Does knowing the attributes improve generation quality?}
\vpara{Does Attribute Fusion improve generation quality? }
Table~\ref{tab:bleu_condition} shows the BLEU scores of the models.
As shown in Table~\ref{tab:bleu_condition}, Attribute Fusion brings an improvement in BLEU score. Attr-A (Both) outperforms the baseline with an advantage of +0.7 BLEU (relatively 9.7\%).
This gain in BLEU score is notable, considering the attributes are categorical and contain little information highly relevant to the product.\footnote{Note that these attributes act as an oracle. 
% as they may derive from the ground truth description in our experiments. 
This simulates the deployed setting where attributes are specified according to user preferences and categories.}

% We also verify that the conditioned model is more effective than the dedicated baseline, as it utilizes the whole dataset, while each dedicated model only sees a fraction of the data.

% \vpara{Does incorporating knowledge improve generation quality?}

% \vpara{Does conditioning on the attributes and incorporating knowledge improve generation diversity?}
\vpara{Do Attribute Fusion and Knowledge Incorporation improve generation diversity? }
Table~\ref{tab:diversity_condition} and Table~\ref{tab:diversity_fact} show the diversity scores of the models, including the ones conditioned on attributes and knowledge.
Both aspect and user category improve the diversity significantly by 46.8\%.
Incorporating knowledge improves the diversity score by 56.4\%.
The improvement in diversity demonstrates that the descriptions generated by our model contain more content than those of the baseline.

\vpara{Does our model capture characteristics of different attributes? }
Table~\ref{tab:accuracy} shows that the accuracy of the attribute classifier, whose computation of the accuracy is described in Equation~\ref{eq:accuracy}.
The low accuracy for the baseline indicates that it is highly possible that the description of the baseline does not match the interested user group.
On the contrary, \kobe\ generates descriptions that better reflect the intended attributes.
\kobe\ obtains a 13.4\% absolute gain in the accuracy for the attribute ``user category'', and a 12.4\% gain in the accuracy for the attribute ``aspect''.
% For the aspect attribute, the gain is 12.4\%.
This means that our model targets the intended user groups more accurately and it is more focused on a specific aspect.

\vpara{Ablation Study }
In Table~\ref{tab:ablation}, we report the results of an ablation study of \kobe. This ablation study is helpful for understanding the impacts of different components on the overall performance.
The components are 1) knowledge, referring to Knowledge Incorporation; 2) user category, referring to the fusion of user category; 3) aspect, referring to the fusion of aspect.
We remove one component successively from \kobe\ until all are removed. Removing knowledge increases the BLEU score from 7.6 to 8.2 but decreases the diversity from 15.1 to 11.1. The external knowledge that is not highly relevant to the product can harm the model's performance in BLEU. Yet it can significantly bring more contents and thus increase the diversity.
Removing user category decreases BLEU sharply from 8.2 to 7.6, and still significantly decreases the diversity from 11.1 to 9.6.
Removing aspect still causes a decrease in BLEU, from 7.6 to 7.2, and a decrease in diversity, from 9.6 to 7.8. These results demonstrate that the attributes do not negatively affect the model's performance in BLEU but make an important contribution to the diversity. In brief, the three components jointly bring a significant improvement to the informativeness of \kobe's generation without sacrificing the performance in BLEU.
% We find that \kobe, with all the components, performs the best. 
% While the diversity is mostly improved by incorporating knowledge, the generation quality is most improved by conditioning on the two attributes.

\begin{table}[t]
    \centering
    \caption{Test BLEU score with different methods of Attribute Fusion.}
    \begin{tabular}{lrrr}
        \toprule
        Model    & Aspect       & User Category   & Both         \\
        \midrule
        Baseline & 7.2          & 7.2          & 7.2          \\
        % (7.32*2797+10.43*545+7.06*1894)/(2797+545+1894)
        Attr-D   & 7.5          & -            & -            \\
        Attr-S   & 7.6          & 7.3          & 7.9 \\
        Attr-T   & 7.3          & 7.1          & 7.7          \\
        Attr-A   & \textbf{7.6} & \textbf{7.5} & \textbf{8.2}          \\
        \bottomrule
    \end{tabular}
    \label{tab:bleu_condition}
\end{table}

\begin{table}[t]
    \centering
    \caption{Test n-gram diversity score with different methods of Attribte Fusion.}
    \begin{tabular}{lrrr}
        \toprule
        Model           & n=3 ($\times 10^{5}$) & n=4 ($\times 10^{5}$) & n=5 ($\times 10^{6}$) \\
        \midrule
        Baseline        & 2.4                   & 7.8                   & 2.0                   \\
        % Attr-D   & -                  & -                  & -                  \\
        Attr-S (Aspect) & 2.8                   & 9.7                   & 2.5                   \\
        Attr-T (Aspect) & 2.8                   & 9.6                   & 2.5                   \\
        Attr-A (Aspect) & 2.8                   & 9.6                   & 2.5                   \\
        Attr-S (User)   & 2.8                   & 9.6                   & 2.5                   \\
        Attr-T (User)   & 2.6                   & 9.1                   & 2.4                   \\
        Attr-A (User)   & 2.7                   & 9.4                   & 2.4                   \\
        Attr-S (Both)   & 3.1                   & \textbf{11.1}         & 2.9                   \\
        Attr-T (Both)   & 2.8                   & 9.7                   & 2.5                   \\
        Attr-A (Both)   & \textbf{3.3}          & \textbf{11.1}         & \textbf{3.0}          \\
        \bottomrule
    \end{tabular}
    \label{tab:diversity_condition}
\end{table}

\begin{table}[t]
    \centering
    \caption{Test n-gram diversity score improved by Knowledge Incorporation.}
    \begin{tabular}{lrrr}
        \toprule
        Model        & n=3 $(\times 10^5)$ & n=4 $(\times 10^5)$ & n=5 $(\times 10^6)$ \\
        \midrule
        Baseline     & 2.4  &   7.8 & 2.0    \\
        % Know-UniAttn & 7.2  &   11.8        \\
        % Know-BiAttn  & 6.8  &   \textbf{12.2}        \\
        Know-BiAttn  &  \textbf{3.1} & \textbf{12.1}  & \textbf{3.3}      \\
        \bottomrule
    \end{tabular}
    \label{tab:diversity_fact}
\end{table}

\begin{table}[t]
    \centering
    \caption{Attribute capturing score with different methods of Attribute Fusion.}
    \begin{tabular}{lrrr}
        \toprule
        Model    & Aspect (\%) & User Category (\%) \\
        \midrule
        Baseline & 63.0        & 71.9            \\
        % (0.956*545+0.909*1894+0.773*2797)/(545+1894+2797)
        % Attr-D   & \textbf{84.1}        & -               \\
        Attr-S (Both)   & 74.0        & \textbf{86.3}            \\
        Attr-T (Both)  & 72.8        & 83.3            \\
        Attr-A (Both)  & \textbf{74.6}        & 86.0            \\
        \bottomrule
    \end{tabular}
    \label{tab:accuracy}
\end{table}

\begin{table}[t]
    \centering
    \caption{Model ablations on 3 model components.}
    \begin{tabular}{lrrr}
        \toprule
        Model           & BLEU & Diversity (n=4) $(\times 10^6)$\\
        \midrule
        \kobe           & 7.6     & \textbf{15.1}         \\
        - knowledge     & \textbf{8.2}     & 11.1         \\
        - user category & 7.6     & 9.6          \\
        - aspect        & 7.2     & 7.8          \\
        \bottomrule
    \end{tabular}
    \label{tab:ablation}
\end{table}

\begin{table}[t]
    \centering
    \caption{The gain from the proposed model \kobe\ over the baseline evaluated by pairwise human judgments.}
    \begin{tabular}{lrrr}
        \toprule
        Model           & Fluency & Diversity & Overall Quality \\
        \midrule
        Baseline        & 3.78     &  3.73 & 3.67        \\
        \kobe           & \textbf{3.95}     &  \textbf{3.79} & \textbf{3.80}        \\
        \bottomrule
    \end{tabular}
    \label{tab:human}
\end{table}

\vpara{Human Evaluation }
We conduct a human evaluation on a randomly sampled subset of the test set.
% To be specific, we randomly select 100 instances to build the set for the evaluation.
Each instance contains an input product title, a description generated by the baseline and a description generated by \kobe.
The selected instances are distributed to human annotators with no prior knowledge about the systems of the descriptions.
Following \citet{jiwei_adversarial}, we require them to independently score the descriptions by three criteria, fluency, diversity and overall quality.
The score of fluency reflects how fluent the description is.
The score of diversity reflects how diversified the description is and whether it contains much competitive content. The score of overall quality reflects whether the description is reasonable, or to say, whether it is consistent with world knowledge. The score range is from 1 to 5. The higher the score is, the more consistent with the criterion the description is. The results are shown in Table~\ref{tab:human}. Consistent with the results of the automatic evaluation of diversity, \kobe\ outperforms the baseline by +0.06 in terms of diversity in human evaluation. However, the improvement in diversity does not sacrifice fluency. Instead, \kobe\ significantly outperforms the baseline by +0.17 in terms of fluency. As to the overall quality, \kobe\ still demonstrates a clear advantage over the baseline by +0.13, showing that our model is able to generate more reasonable descriptions with the introduction of external knowledge.

\subsection{Case Studies}

In this section, we perform case studies to observe how our proposed methods influence the generation so that the model can generate different contents based on different conditions.

Table~\ref{tab:varying_aspects} presents the generated examples by the models with different aspects.
We compare the description generated by models conditioned on two aspects, ``aspect'' and ``function''.
The product description in the left column contains the content about the appearance of the product, such as the description of the pattern of lotus, but the product description in the right column focuses more on the functions with some expressions about its functional features, such as ``transparency'', ``light'', etc. 

We also observe that it can also generate different descriptions based on different user categories. It has the potential to generate texts of multiple styles. In Table~\ref{tab:varying_users}, it can be found that the two user categories, representing two extremely different styles, can make a difference in the generation.
With the category ``housewife'', the generated description focuses on the quality and safety of the product.
But with the category ``geek'', it describes the same desk as a ``computer'' desk and focuses more on its functional features.

% \vpara{Demonstration video }
% We show many more deployed examples of product descriptions generated by our framework in a demonstration video, which is available online at \href{https://youtu.be/WXhP0NcxZ6E}{\bf https://youtu.be/WXhP0NcxZ6E}.

% We focus on two tags, ``housewife'' and ``geek''. In our common sense, the text style suitable for housewives should be more about family, especially children, while the text style for geek should focus on technologies and functions.
% The results are placed in the left and right column respectively.
% In the left column, we can see an emphasis on the appearance of the product while in the right column, the description mainly focuses on the functions.

% \begin{CJK*}{UTF8}{gbsn}
% % 我觉得 appearance 和 function 这边的第一个（柳条编制）和第四个（台灯）两个例子相对更好一些

% 我们生成的描述尽管仍然以类似的比较宽泛的话作为开头，但渐渐开始进行了不同方面的 emphasis。依次指定 aspect 以生成全面描述，感觉不太好说怎么去处理重复的内容，感觉可以再吹一下 diversity？

% % 感觉后面第三个例子最好
% 我们也尝试了对同一个商品改变输入的用户标签。在第三个例子中，给定一个 basic information 是 ``书桌书架组合家用实欧实木电视柜书桌一体简约写字桌儿童学习桌''的商品。当我们目标用户是家庭主妇时，生成的描述“儿童书桌”有这种面向家庭的；当目标用户是“数码达人”时，强调了“电脑桌”、“收纳需求”等，说明我们的方法能根据接受者产生适合的文本。

% 第二个儿童运动鞋变成帅气迷人有点奇怪，不过也能说明我们的方法的有效性。

% 因为主要是对文本进行分析，感觉做到更 deep 需要一些电子商务的专业知识。。感觉如果只是简单的对比是不是一般都写在 Supplementary Material 里。。

% \end{CJK*}

% Jie: Give qualitative results analysis. Go deep.

\begin{CJK*}{UTF8}{gbsn}
    \begin{table*}[t]
        \centering
        \caption{Each pair of descriptions is generated by varying the aspect attribute while fixing the product title as input.}
        \small
        \begin{tabularx}{0.96\textwidth}{XX}
            \toprule
            \textbf{Varying product aspects}                                                                                                                                                                \\
            % Product                  & 柳编野餐篮藤编保温篮野餐篮子藤编手提折叠收纳篮野餐包野营                                                                             \\
            \textit{``appearance''}                                                                                                                                                     & \textit{``function''} \\
            \midrule
            The Chinese-style lamp is made of superior resin. It is healthy, environmentally-friendly. The hand-carved pattern of lotus is lifelike and full of Chinese-style elements. &
            % 新中式风格的台灯，采用优质的树脂材质制作而成，环保健康，经久耐用，透光性好，光线柔和不刺眼，适合多种场合使用。
            The Chinese-style lamp is made of superior resin. It is healthy, environmentally-friendly, durable. It is also with good transparency and its light is soft, which is suitable for many situations. \\
            \bottomrule
        \end{tabularx}
        \label{tab:varying_aspects}
    \end{table*}
\end{CJK*}

\begin{CJK*}{UTF8}{gbsn}
    \begin{table*}[t]
        \centering
        \caption{Each pair of descriptions is generated by varying the user category attribute while fixing the product title as input.}
        \small
        \begin{tabularx}{0.96\textwidth}{XX}
            \toprule
            \textbf{Varying user categories}                                                                                                                                                                                                                                                                                                                   \\
            \textit{``housewife''}                                                                                                                                                                                                                                                                                                       & \textit{``geek''}   \\
            \midrule
            % 书桌书架组合家用实欧实木电视柜书桌一体简约写字桌儿童学习桌 Combination of desk and shelf, Euro-style wooden desk
            % 欧式风格的一款儿童书桌，精选优质的实木材质打造而成，质地坚硬，结构稳固，经久耐用，环保的油漆喷涂，健康安全。
            The Euro-style children's desk is made of superior wood, which is hard, solid and durable. It is covered with environmentally friendly oil paint, which is healthy and safe.                                                                                                                                                 &
            % 欧式实木电脑桌，精选优质橡胶木打造，木质坚硬，纹理清晰，结构稳固，环保油漆，安全健康，多功能储物设计，满足您的收纳需求。
            The Euro-style wooden computer desk is made of superior rubberwood, which is hard. It has a clear texture and solid structure and it is covered with environmentally friendly oil paint, which is healthy and safe. It has a multifunctional design for storage, which can meet your requirements for gathering.                                   \\
            \bottomrule
        \end{tabularx}
        \label{tab:varying_users}
    \end{table*}
\end{CJK*}

% !TEX root = ./0.main.tex

\section{Related Work}

%In recent years, 
Neural networks have 
%dominated the study of 
been widely used for automatic text generation. 
In this section, we 
%focus on this field and 
briefly review related literature. 

\vpara{Text generation}
A basic model for text generation is the attention-based Seq2Seq model \citep{seq2seq, attention, stanford_attention}.
% sequence-to-sequence (Seq2Seq) learning, which is based on the encoder-decoder framework \citep{seq2seq, GRU}. Furthermore, with the success in neural machine translation \citep{attention, stanford_attention}, the attention mechanism has become a basic component in the framework \citep{attention, stanford_attention}. 
The attention-based Seq2Seq model has demonstrated to be effective in a number of tasks of text generation, including neural machine translation \citep{seq2seq, attention, GNMT}, abstractive text summarization \citep{abs, chopra, IBM_sum}, dialogue generation \citep{vinyals2015neural, serban_dialog, bordes}, etc. 
% Besides, some efforts have been also made to leverage generative adversarial networks (GANs) and reinforcement learning (RL) to enhance the quality of generation
% through a sophisticated design of model and rewards 
% \citep{jiwei_adversarial, jiwei_rl, dp-gan}. 
Generally speaking, the Seq2Seq model has become one of the most common frameworks for text generation.

While most of the Seq2Seq models are based on RNN, recently researchers have proposed frameworks based on CNN \citep{fairseq} and attention mechanism \citep{transformer}. The Transformer has achieved state-of-the-art results in neural machine translation and rapidly become a popular framework for sequence-to-sequence learning due to its outstanding performance and high efficiency. \citep{BERT} proposed BERT, a pretrained language model based on the Transformer, has achieved the state-of-the-art performances on 11 natural language processing tasks. Following the studies, we implement our methods upon the Transformer framework.

\vpara{Product description generation} As for product description generation, previous studies focused on statistical frameworks such as \citep{wang2017statistical}, which incorporates statistical methods with the template for the generation of product descriptions. 
%We consider that such generation can be highly restricted 
\citet{gerani2014abstractive} also generates summarization of product reviews by applying a template-based NLG framework.
Such methods are limited by the hand-crafted templates. 
To alleviate the limitation, researchers adopted deep learning models and introduce diverse conditions to the generation model.
\citet{lipton2015capturing} proposed a method to generate reviews based on the conditions of semantic information and sentiment with a language model, and \citet{tang2016context} conditioned their generation on discrete and continuous information. Related literature can be also found in 
%approaches have been adopted in a number of models for stylistic and affective generation 
\citep{li2016persona, asghar2018affective, herzig2017neural, hu2017toward, ficler2017controlling}.
%and move to the automatic generation of product description, which, 
To the best of our knowledge, our research takes the first attempt 
%is the first work that focuses on 
to use neural methods and sequence-to-sequence learning for product description generation by considering personalization and informativeness.
%. Furthermore, as our objective is to generate personalized and informative product description, the effective incorporation of specified conditions is significant. 

% \noindent\textbf{Sequence to Sequence Learning}

% \reminder{paraphrase: In such frameworks, the encoder takes the input sequence and encode the sequence into a fixed-length vector as a latent representation. Following the attention mechanism \cite{attention} which is the mechanism for decoding the latent representation to a target sequence, the NMT framework is built upon the maximum likelihood estimation (MLE) for training}.

% \noindent\textbf{Conditioned Language Models}
% \reminder{paraphrase: \cite{lipton2015capturing} use character-level rnns conditioned on semantic information and sentiment, to generate product reviews, while \cite{tang2016context} generate such reviews using an lstm conditioned on input ‘contexts’, where contexts incorporate both discrete (user, location etc) and continuous information.}
% Similar approaches have been adopted in a number of models for stylistic and affective generation \citep{li2016persona, asghar2018affective, herzig2017neural, hu2017toward, ficler2017controlling}.

% \noindent\textbf{Knowledge Grounded Text Generation Models}
% Incorporating unstructured texts \cite{ghazvininejad_knowledge-grounded_2017, long2017knowledge, lowe2015incorporating}

% \noindent\textbf{Product Description Generation}

% Template-based methods. Statistical methods \citep{wang2017statistical}.

% !TEX root = ./0.main.tex

\section{Conclusion}

In this paper, we study an interesting problem on how to automatically generate personalized product descriptions in an e-commerce platform.
%by considering personalization and informativeness.
%incorporate neural methods with
% so that such product description can enhance the quality of online recommendation and impose significant influence upon the users' decision making. The research is conducted on the data of Taobao, and we propose \textit{TaoDescribe} a large-volume dataset for product description generation, which is retrieved from the real-world data in Taobao. Furthermore, motivated by the recent advances in text generation, we focus on generating personalized and informative product description. 
We propose a novel model based on the Transformer framework which incorporates multiple types of information efficiently, including product aspects, user categories, as well as the knowledge base. Our extensive experiments show that our method outperforms the baseline models
through a series of evaluations, including automatic evaluation and human evaluation.
%, showing advantages of coherence, diversity and relevance. Besides, o
We have successfully deployed the proposed framework onto Taobao, one of the world's largest online e-commerce platforms. A large volume dataset for product description generation, namely \textbf{TaoDescribe}, has been generated and annotated, which can be used as a benchmark dataset for future research in this field.
 
%The research is conducted on the data of Taobao, and we propose \textit{TaoDescribe} a large-volume dataset for product description generation, which is retrieved from the real-world data in Taobao.
%Our analyses also show that the incorporation of the introduced information is able to modify the generated contents from the specified perspective, making such generation more interpretative. 
%Be there as it may,  there is still room for the relevance of generation to the input product information to be enhanced and the integration of external commonsense, which are important for future work.

\begin{acks}
The work is supported by the
NSFC for Distinguished Young Scholar 
%jie tang
(61825602),
%chunyuan zhou
NSFC (61836013), 
and a research fund supported by Alibaba.

\end{acks}

\bibliographystyle{ACM-Reference-Format}
\bibliography{reference}

\appendix
% !TEX root = ./0.main.tex

% This supplement can only be used to include (i) information necessary for reproducing the experimental results, insights, or conclusions reported in the paper (e.g., various algorithmic and model parameters and configurations, hyper-parameter search spaces, details related to dataset filtering and train/test splits, software versions, detailed hardware configuration, etc.), and (ii) any pseudo-code, or proofs that due to space limitations, could not be included in the main nine-page manuscript, but that help in reproducibility (see reproducibility policy below for more details).

% Authors are strongly encouraged to make their code and data publicly available whenever possible. In addition, authors are strongly encouraged to also report, whenever possible, results for their methods on publicly available datasets. Algorithms and resources used in a paper should be described as completely as possible to allow reproducibility. This includes experimental methodology, empirical evaluations, and results. The authors are encouraged to take advantage of the optional two-page supplement to provide the appropriate information. The reproducibility factor will play an important role in the assessment of each submission.

\section{Appendix}
\label{sec:appendix}

\subsection{Details on the Experimental Setup and Hyperparameters}

\vpara{Hyper-parameter Configurations }
We use a six-layer Transformer Encoder for both the title encoder and the knowledge encoder and a two-layer Transformer Decoder for the description decoder.
All input embedding dimensions and hidden sizes are set to $d_{model}=512$.
Following \cite{transformer}, we empirically set the size of the Transformer FFN inner representation size ($d_{ff}=2048$) to four times of the embedding size ($d_{model}=512$).
We use ReLU \citep{ReLU} as the activation function for all models.
% Except for the two main components, the other units in an encoder layer include residual connection \citep{resnet} and layer normalization \citep{LayerNorm}.
During training, the batch size is set to 64.
We use Adam optimizer \citep{adam}  with the setting $\beta _ { 1 } = 0.9 , \beta _ { 2 } = 0.998 \text { and } \epsilon = 1 \times 10 ^ { - 9 }$.
The learning rate is set to $1 \times 10 ^ { - 4 }$.
Gradient clipping is applied with range $\left[-1, 1\right]$.
We also applied a dropout rate of 0.1 as regularization.
At the inference stage, we use beam search with a beam size of 10.

\vpara{Hardware Configuration }
The experiments are conducted on a Linux server equipped with an Intel(R) Xeon(R) Platinum 8163 CPU @ 2.50GHz, 512GB RAM and 8 NVIDIA V100-SXM2-16GB GPUs.
Our full set of experiments took about 3 days with this multi-GPU setting.
We expect that a consumer-grade single-GPU machine (e.g., with a Titan X GPU) could complete our main experiments, including \kobe\ and the baseline, in 4 days.

\vpara{Software }
All models are implemented in PyTorch \citep{paszke2017automatic} version 0.4.1 and Python 3.6.
We'll soon release our code for preprocessing the dataset and running all the experiments.

\subsection{Additional Details on the Dataset }

In this section, we provide some additional, relevant dataset details.
Our code and the \textbf{TaoDescribe} dataset are available at \textbf{\url{https://github.com/qibinc/KOBE}}.

\vpara{Data Preprocessing }
We observe that the samples are noisier if the product title or the product description is too long.
For this reason, we discarded pairs with a title longer than 100 tokens or with a description longer than 150 tokens.
We then substituted tokens that appeared less than 5 times in our dataset with the <UNK> token.
Furthermore, a product title $\boldsymbol{x}$ in our dataset may be coupled with multiple descriptions $\boldsymbol{y}$ (e.g. covering different aspects, targeting different users, or written by different people).
We split our dataset by products to keep product titles in the test set from being seen during training.
The dataset is finally split into a training set with 2,114,034 instances, a validation set with 9,917 instances and a test set with 5,236 instances.
The vocabulary size is 5,428 for product titles and 9,917 for descriptions.
The average lengths of product titles and descriptions are 31.4 and 90.2 tokens, respectively.

\vpara{Aspects Selection and Annotation}
Based on the statistics of the dataset and empirical support by domain experts, we set the number of product aspects $|\mathcal{A}_1|$ to three.
We introduce the heuristic method used to extract the aspects from the description in detail and demonstrate its effectiveness as follows.

\begin{enumerate}
    \item We extract all the descriptions for the products in the dataset and run word2vec \citep{mikolov2013distributed} on the description sentences to obtain the embedding for each word in the description vocabulary.
    \item For each adjective in the description vocabulary,\footnote{We use THULAC \citep{sun2016thulac} for part-of-speech tagging.}  we compute the cosine distance between its embedding and the embedding of each of the $|\mathcal{A}_1|$ aspect words respectively and obtain similarity scores $\left\{s_1, s_2, \ldots, s_{|\mathcal{A}_1|}\right\}$.
    \item We discard the adjective if $\max\left\{s_1, s_2, \ldots, s_{|\mathcal{A}_1|}\right\} < \gamma {\sum_{k=1}^{|\mathcal{A}_1|} s_k}$,\footnote{We empirically set $\gamma$ to 0.8.} which means the adjective cannot be categorized to any aspect. Each remaining adjective in the description vocabulary is then assigned to the aspect with the highest similarity score.
    \item We then automatically annotate each description $\boldsymbol{y}\in\mathcal{Y}$ in the dataset with an aspect $a_1$, based on the semantic distance between the description and the set of adjectives belonging to the aspect.
\end{enumerate}

\begin{figure}
    \centering
    \includegraphics[width=0.45\textwidth]{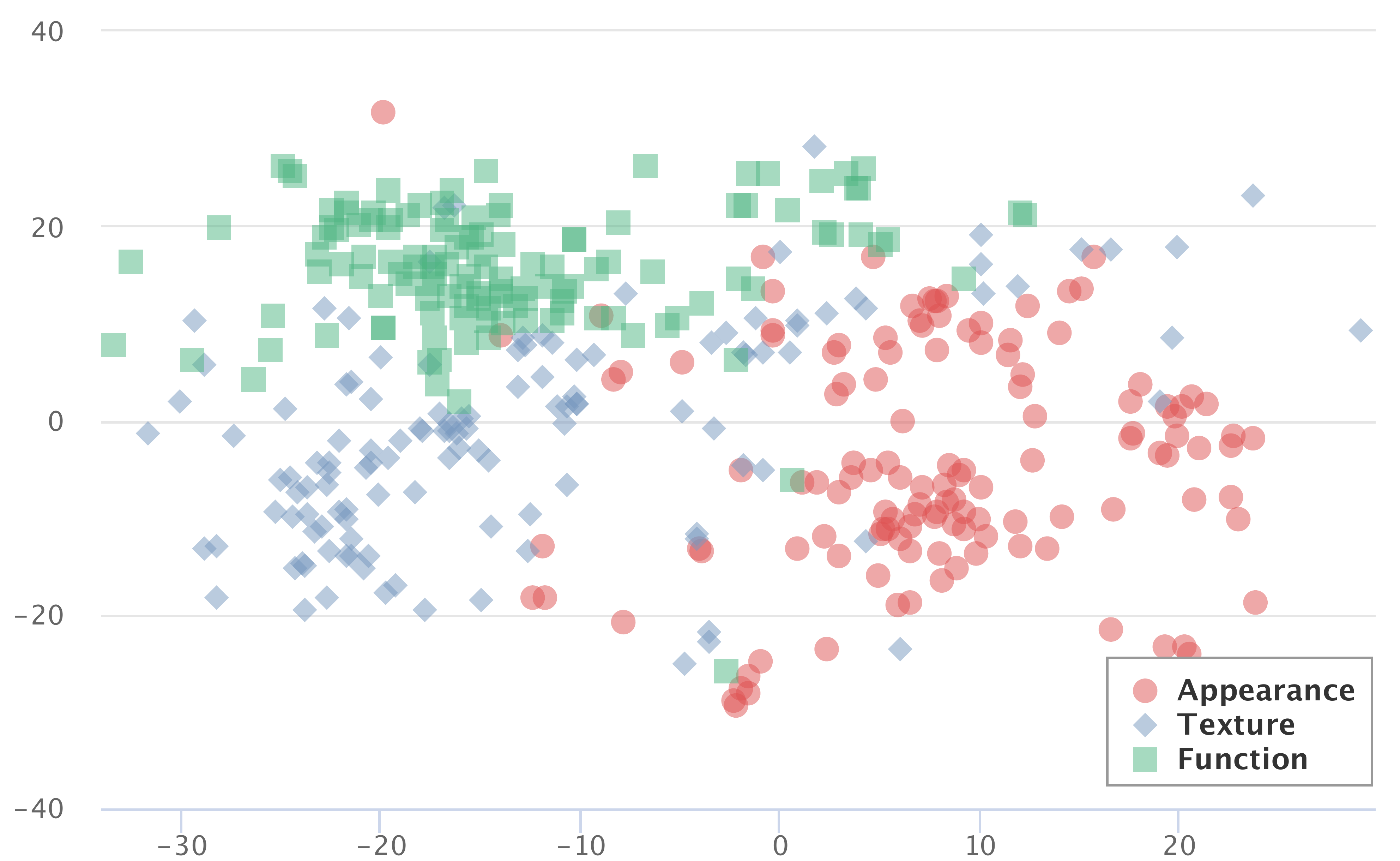}
    \caption{Word embeddings of adjectives in the three aspects projected onto a 2D plane by t-SNE \citep{maaten2008visualizing}. Each color represents an aspect and each point represents an adjective in the description vocabulary.}
    \label{fig:aspect_vis}
\end{figure}

We give examples and the number of obtained adjectives for the three aspects in Table~\ref{tab:aspects}.
Furthermore, we visualize the word embeddings of adjectives in the three aspects in Figure~\ref{fig:aspect_vis}.\footnote{128 adjectives for each aspect are displayed for clear visualization.}
As shown in the figure, adjectives belonging to different aspects are mostly well separated.
This means different aspects can be distinguished in the semantic space and further supports our selection of the three aspects.

\begin{table}
    \centering
    \caption{Examples of adjectives in different product aspects.}
    \begin{tabular}{lrr}
        \toprule
        \textbf{Aspect} & \textbf{Num.} & \textbf{Examples}                 \\
        \midrule
        appearance      & 754           & contracted, elegant, monochrome   \\
        texture         & 235           & soft, gentle, smooth, comfortable \\
        function        & 265           & household, convenient, automatic  \\
        \bottomrule
    \end{tabular}
    \label{tab:aspects}
\end{table}

\vpara{User Categories Collection}
In our platform, there are 41 predefined ``interest tags'' based on expert knowledge and statistics of user behaviors.
These interest tags are mapped to ``product categories''.
Each user is then soft assigned to these tags, based on his/her browsing, favoring and purchasing history on different product categories.
Each description can then be assigned to a user category, according to the major group of users that have clicked or stayed long enough on this description.
Ideally, the number of user categories should be the same as the number of interest tags.
In practice, we find that user categories with a low appearing frequency can cause some noise during training, for there are too few corresponding descriptions.
To solve this problem, we replace user categories appearing less than 5 times with the <UNK> token.
Finally, the number of obtained user categories $|\mathcal{A}_2|$ is 24.
Table~\ref{tab:user_categories} lists examples of user categories and their frequency in $\mathcal{Z}$.

\begin{table}[t]
    \centering
    \caption{Examples of user categories.}
    \begin{tabular}{lrr}
        \toprule
        \textbf{User Category} & \textbf{Frequency} \\
        \midrule
        fashion girl           & 34,916              \\
        housewife              & 29,626              \\
        foodie                 & 17,049              \\
        geek                   & 15,882              \\
        ...                    & ...                \\
        \bottomrule
    \end{tabular}
    \label{tab:user_categories}
\end{table}

\vpara{User Categories Annotation}
To overcome the problem that user feedbacks on $\mathcal{Y}$ are very sparse, we collect another dataset $\mathcal{Z}$ and train a user category classifier on $\mathcal{Z}$ to annotate $\mathcal{Y}$.
% We first introduce the details about $\mathcal{Z}$.
In addition to $\mathcal{Y}$ which only contains single product descriptions, we make $\mathcal{Z}$ to incorporate descriptions of products and user feedbacks.
We then discard descriptions with the number of user feedbacks (clicks) less than 100 and obtain 143,154 descriptions and their corresponding user categories.
We find that this increase in data size significantly increases the generalization ability of the trained classifier, despite the slight difference between $\mathcal{Z}$ and $\mathcal{Y}$.

Now we introduce the architecture and training procedure of the user category classifier $M$.
Due to the massive scale of our dataset, the classifier used to annotate the user category for the descriptions should be both effective and efficient.
We applied fastText \citep{joulin2016bag} as the text classifier but soon find that it is prone to overfitting.
The performance heavily relies on carefully setting the number of training epochs to stop early.
For this reason, we adopt convolutional neural networks (CNN) for sentence classification \citep{kim2014convolutional} as the user category classifier $M$.
We clarify the architecture and configuration of $M$ as follows.

Given the input sentence $\boldsymbol{z} = (z_1, z_2, \ldots, z_n)$, we first embed each word into $d$ dimensions and obtain $\boldsymbol{e} = (e_1, e_2, \ldots, e_n), \boldsymbol{e} \in \mathbb{R}^{h\times d}$.
Then, a feature $c_i$ is generated by applying a filter $\boldsymbol { w } \in \mathbb { R } ^ { h \times d }$ on a window of $h$ words $\boldsymbol { e } _ { i : i + h - 1 }$.
% $$c _ { i } = f \left( \boldsymbol { w } \cdot \boldsymbol { e } _ { i : i + h - 1 } + b \right),$$
% where $\boldsymbol { w } \in \mathbb { R } ^ { h \times d }$ is a filter,
% $b \in \mathbb { R }$ is a bias term and $f$ is a non-linear function such as ReLU.
A feature map $\boldsymbol { c } = \left[ c _ { 1 } , c _ { 2 } , \ldots , c _ { n - h + 1 } \right]$ is then obtained by applying the filter $\boldsymbol{w}$ to each possible window of words in the sentence $\left\{ \boldsymbol { e } _ { 1 : h } , \boldsymbol { e } _ { 2 : h + 1 } , \dots , \boldsymbol { e } _ { n - h + 1 : n } \right\}$.
The feature $\hat{c}$ corresponding to this particular filter is finally obtained by a max-pooling operation over the feature map $\hat { c } = \max \{ \boldsymbol { c } \}$.
Finally, features computed by convolutional filters are projected to the user category space through a fully connected layer.
The model is then trained by maximizing the likelihood $P(a|\boldsymbol{z}; \theta_M)$ of the correct user category $a$, where $\theta_M$ denotes the parameters of the classifier $M$.

For the classifier, we use word embedding size $d=100$ and apply three types of filters with window sizes 3, 4 and 5, each with 100 filters.
During training, the batch size is set to 64.
We apply the cross entropy loss and an Adam optimizer with the default setting $\beta _ { 1 } = 0.9 , \beta _ { 2 } = 0.999 \text { and } \epsilon = 1 \times 10 ^ { - 8 }$.
The learning rate is set to $1 \times 10 ^ { - 3 }$.
We also applied a dropout rate of 0.5 to prevent overfitting.
The testing classification accuracy for 24 user categories on $\mathcal{Z}$ is 81.0\%.
Finally, the classifier $M$ trained on $\mathcal{Z}$ is used to annotate the user category for each description $y\in \mathcal{Y}$.
We also provide $\mathcal{Z}$ along with the dataset for full reproducibility.

\end{document}